\documentclass[10pt,twocolumn,letterpaper]{article}

\usepackage{cvpr}
\usepackage{times}
\usepackage{epsfig}
\usepackage{graphicx}
\usepackage{amsmath}
\usepackage{amssymb}

\usepackage[breaklinks=true,bookmarks=false]{hyperref}

\cvprfinalcopy 

\def\cvprPaperID{****} 

\begin{document}

\title{Let's Transfer Transformations of Shared Semantic Representations}

\author{
    Nam Vo, Lu Jiang, James Hays \\ 
    Georgia Tech, Google AI \\
    {\tt\small namvo@gatech.edu, lujiang@google.com, hays@gatech.edu}
}

\maketitle

\begin{abstract}
With a good image understanding capability, can we manipulate the images high level semantic representation? Such transformation operation can be used to generate or retrieve  similar images but with a desired modification (for example changing beach background to street background); similar ability has been demonstrated in zero shot learning, attribute composition and attribute manipulation image search. In this work we show how one can learn transformations with no training examples by learning them on another domain and then transfer to the target domain. This is feasible if: first, transformation training data is more accessible in the other domain and second, both domains share similar semantics such that one can learn transformations in a shared embedding space. We demonstrate this on an image retrieval task where search query is an image, plus an additional transformation specification (for example: search for images similar to this one but background is a street instead of a beach). In one experiment, we transfer transformation from synthesized 2D blobs image to 3D rendered image, and in the other, we transfer from text domain to natural image domain.

\end{abstract}

\section{Introduction}

\begin{figure}[t]
\begin{center}
\includegraphics[width=0.95\linewidth]{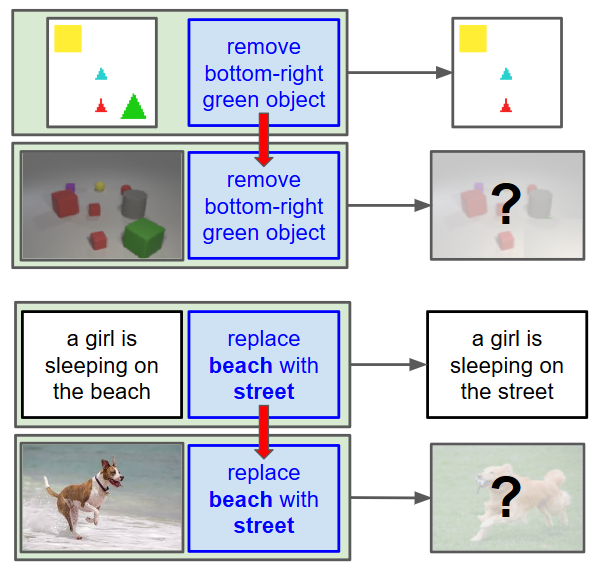}
\end{center}
   \caption{Example of transformations that can be applied to 2 different domains. If they share similar semantics, we can learn on one and transfer to the other.}
\label{fig:intro}
\end{figure}

A smart image to image retrieval system should be able to incorporates user feedbacks such as relevance \cite{rui1998relevance}, attribute \cite{kovashka2012whittlesearch,zhao2017memory}, spatial layout \cite{mai2017spatial} or text \cite{guo2018dialog,our}. This paper studies the above application; the scenario is that user want to search for images that similar to a reference image, but with an additional specification (such as ``change object color to red" or ``switch beach background to street"). We formulate a function parameterized by the specification, taking the reference image feature and outputting a new one that represents what the user is looking for; in this work we call such function ``transformation".

Training a vision system that can perform such kind of semantic manipulation can be straightforward. That is if there's enough labeled data, which unfortunately is not always the case: finding images which contains desired transformation might not be possible, manually transform the images in its native domain could be a costly annotation. In this work we explore an alternative: learn the transformation function in another domain that shares similar semantics. It could be a totally different domain, or a customized simplified version of the original domain.

There are many use cases in which collecting examples in one domain is much easier, or cheaper:
\begin{itemize}
\item We demonstrate this on the synthesized dataset CSS \cite{our}, where the same scene can be rendered 3D realistically or 2D simplistically. Rendering these scenes in 3D even with a GPU is still multiple magnitudes slower.
\item The second use case is image and caption \cite{lin2014microsoft,young2014image,wang2016learning}. Editing and manipulate images are highly specialized skills while manipulating text is the first thing people learn in school. In fact in our experiment we show how to generate ``word replacing" transformation automatically on the fly for training.
\item Other scenarios includes 3D shape and caption \cite{chen2018text2shape}, streetview image, computer generated image and corresponding category segmentation map \cite{cordts2016cityscapes,richter2016playing}, facial images and corresponding facial landmarks \cite{ren2014face}, scene image and scene graph \cite{johnson2018image}, etc. The later domains are usually easier to express transformation on. Even without manual annotation, one can automatically generate ``change trees to buildings" transformation on segmentation map, or ``make mouth smaller" transformation on facial landmarks.

\end{itemize}

In this work, we show that one can learn a transformation in one domain and transfer it to another domain by sharing a joint embedding space, assuming they have similar semantics and the transformation is universal to both domains. We demonstrate its usefulness on the image to image retrieval application, where the query is now a reference image plus an additional specification to enhance the query's expressiveness \cite{our}. 2 datasets are experimented with: the synthesized dataset CSS \cite{our} and the image-caption dataset COCO 2014 \cite{lin2014microsoft}, shown in Figure 1.

\section{Related Works}

\textbf{Image retrieval}:
beside traditional text-to-image \cite{wang2016learning} or image-to-image retrieval \cite{wang2014learning} task, there are many image retrieval applications with 
other types of search query such as: sketch \cite{sangkloy2016sketchy}, scene layout \cite{mai2017spatial}, relevance feedback \cite{rui1998relevance,jiang2012leveraging}, product attribute feedback \cite{kovashka2012whittlesearch,zhao2017memory, han2017automatic, ak2018learning}, dialog interaction \cite{guo2018dialog} and image text combination query \cite{our}. In this work, the image search query will be a combination of a reference image and a transformation specification. In our setup, labeled retrieval examples are not available, hence a standard training procedure like \cite{zhao2017memory,our} does not work.

Perhaps the most relevant approach is to learn a joint embedding space and analyze its structure \cite{han2017automatic, chen2018text2shape}. Here we learn a transformation function and use shared embedding as a way to transfer it.


\textbf{Zero shot learning} aims to recognize novel concepts relying on side data such as attribute \cite{lampert2009learning,akata2013label,deng2014large} or textual description \cite{reed2016learning,frome2013devise}. This side data represents high level semantics with structure and therefore can be manipulated, composed or transformed easily by human. On the other hand, corresponding manipulation, but in the low level feature domain (like raw image), is more difficult.

\textbf{GAN image generation, style transfer or translation } is an active research area where high level semantics modification or synthesization of images is done \cite{reed2016generative,isola2017image,choi2017stargan,sangkloy2016scribbler,xian2016texturegan,raj2018swapnet}. For example, ``style" can represent a high level semantic feature that one want to enforce on the output. Reed et al \cite{reed2016generative} generate images from  reference images with textual description of new ``style". 
Another relevant research area is works on translation between scene image, scene graph and text captions \cite{xu2017scene,johnson2018image}.

\textbf{Embedding and Joint embedding} is one of the main approaches for many retrieval and zero shot learning tasks. It usually relies on metric learning \cite{weinberger2009distance,wang2014learning,vo2018generalization} in retrieval context; though other supervised setting or even unsupervised learning \cite{radford2015unsupervised} can also work. The result will be encoders that embed raw input into a high semantic level feature space, where retrieval or recognition is performed. Our work concerns performing transformation within such space. In \cite{radford2015unsupervised}, it is demonstrated that walking or perform vector arithmetic operation there can translate to similar high level semantic changes in the raw image space.

\textbf{Synthesized data, simulation and domain adaptation:} these areas are at high level similar to what we want to do: perform learning on another domain where label is available and apply to the target domain \cite{richter2016playing,shrivastava2017learning,tzeng2017adversarial}.
Here the source and target domains here are similar and the goal is to finetune the model trained on one domain for another by bridging the gap between 2 domains. Differently the task we are studying here requires transferring between 2 completely different domains (i.e. image and text) and so provides similarity supervision to facilitate that. 

\section{Method}

\begin{figure*}
\begin{center}
\includegraphics[width=0.85\linewidth]{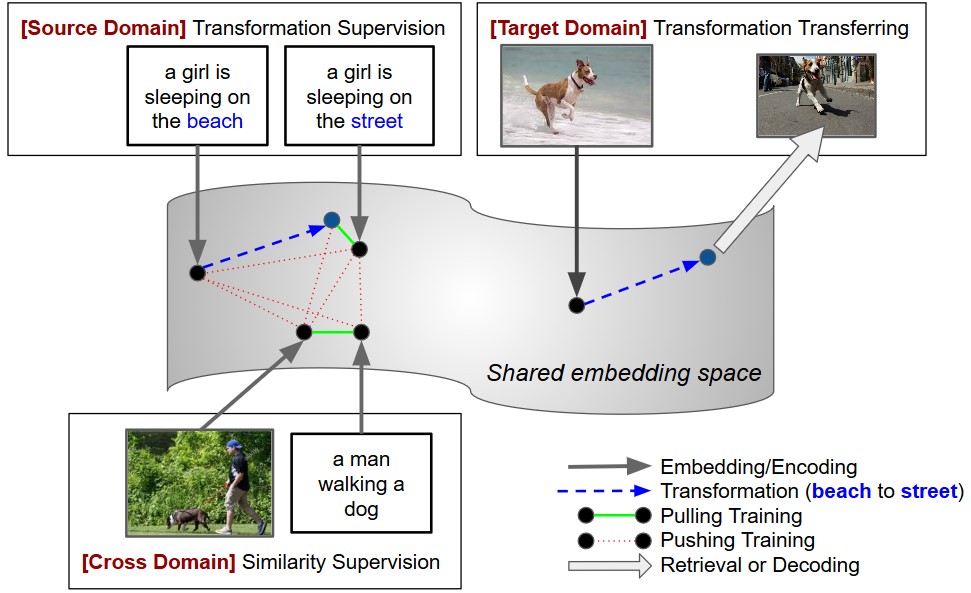}
\end{center}
   \caption{Overview of our transformation transferring by sharing embedding space.}
\label{fig:pipeline}
\end{figure*}

We study the problem of learning a transformation function in one domain and transfer it to another domain; we choose image retrieval for demonstration and quantitative experiments though the formulation might be applicable to other domains or other tasks.

First we formalize the problem: source domain $S$ and the target domain $T$ have the similar underlying semantics; corresponding supervision is the set $E = \{(e^s_i, e^t_i)\}^n_{i=1}$ where $e^s_i \in S$, $e^t_j \in T$ are labeled similar if $i=j$ and non-similar otherwise.

Supervision for transformation is provided for the source domain: $X^S = \{(q^s_i,t_i, r^s_i)\}^m_{i=1}$ where $q^s_i \in S$ is a "before-transformation" example, $t_i$ is transformation specification/parameter and $r^s_i \in S$ is a "after-transformation" example. Note that the set of $e^s$, $q^s$ and $r^s$ can be the same, intersected or mutually exclusive.

Given a similar labeled set but for testing on the target domain instead $X^T = \{(q^t_i,t_i, r^t_i)\}^m_{i=1}$, the task is to for each query $(q^t_i, t_i)$ retrieve the correct $r^t_i$ in the pool of all examples. We propose to learn to do that by: (1) learn a shared semantic representation using supervision $E$ and (2) learn to transform that shared representation using supervision $X^S$.

\subsection{Deep Shared Embedding}

The first step is to learn the embedding functions $f^s_{embed}, f^t_{embed}$ for each domains. For convenience, we denote:
\begin{equation}
\hat{x} = f_{embed}(x) =
\begin{cases} 
 f^s_{embed}(x) \text{ if $x \in S$} \\
 f^t_{embed}(x) \text{ if $x \in T$} \\
\end{cases}
\end{equation}

We will make use of recent advance in deep learning for this task. In particular, CNN will be used as encoder if the domain is image and LSTM will be used if it is text.

The learning objective is for $\hat{e}_i=f_{embed}(e_i)$ and $\hat{e}_j=f_{embed}(e_j)$ to be close each other in this space if $e_i$ and $e_j$ (which can be from the same domain or different) are labeled similar, and far from each other otherwise. 
Any any distance metric learning loss function $f_{metric}$ can be used:

$$ L_{embed} = f_{metric}(\{(\hat{e}_i, \hat{e}_j) | \forall SimilarPair (i,j)\}) $$

We used the following function $f_{metric} = f_{CE}$:
$$ f_{CE}(\{(a_i, b_i)\}_{i=1..N}) = \frac{1}{N} \sum_{i=1}^N CE(scores=p_i, label=i) $$

Where $CE(scores, labels)$ is the softmax cross-entropy function,  
$ p_i = s[a_i^T b_1, a_i^T b_2, ... a_i^T b_N]$.

\subsection{Transformation in the Embedding Space}

The transformation is formalized as a function $ f_{transform}(\hat{e}, \hat{t}\text{ }) $ where $\hat{e}, \hat{t}$ are the feature representation of example $e$ and the transformation $t$ respectively; we extend the definition of $f_{embed}$ so that: $\hat{x} = f_{embed}(x) = f^{t}_{embed}(x)$ when $x$ is transformation specification.

There is many kinds of feature fusion techniques which can be used as the transformation function. For example the simple concatenate fusion that we will use:
$$f_{transform}(\hat{e},\hat{t}\text{ }) = NN(concat(\hat{e}, \hat{t}\text{ }))$$
Where $concat()$ is concatenation operation and $NN()$ is a (learnable) 2 layer feed forward network. For reference, \cite{our} benchmarks different image text fusion mechanisms in image retrieval context, we will also experiment with their proposed method TIRG.

For each transformation example $(q, t, r)$, the learning objective is for $\tilde{r} = f_{transform}(\hat{q},\hat{t}\text{ })$ close to $\hat{r} = f_{embed}(r)$ while being far from other features in the embedding space. We use the same metric learning loss function in previous section to enfore that objective.

$$ L_{transform} = f_{metric}(\{(\tilde{r}, \hat{r}) | \forall (q, t, r)\}) $$

\subsection{Domain-blind Example Augmentation}

\begin{table*}
\begin{center}
\begin{tabular}{|l|ccc|}
\hline
Training approach & 2D-to-2D & 2D-to-3D & 3D-to-3D \\
\hline\hline
2D-to-2D image retrieval training (16k examples) & 73 &  - &  - \\
2D-to-3D image retrieval (16k) & - &  43 &  - \\
3D-to-3D image retrieval (16k) & - & - & 72 \\
2D-to-2D image retrieval (16k) + 2D-3D shared embedding (1k) & 73 & 57 & 71 \\
\hline
\end{tabular}
\end{center}
\caption{R@1 Retrieval performance on CSS.}
\label{tab:css}
\end{table*}

Note that when defining above functions, we remove domain specific notation from the variables. It's so that they are applicable regardless which domain the examples are from. In general $E$ can also include in-domain similarity supervision and $X$ can also include cross-domain transformation supervision, if available.

If the examples in the supervision overlap, transitivity can be applied, for instance if $e_i$ and $e_j$ are labeled similar, and $(e_i, t, r)$ is a valid transformation example, then $(e_j, t, r)$ is also a valid transformation.

If the transformation is reversible (for example [add red cube] and [removed red cube]), then for each $(q, t, r)$, we have $(r, reverse(t), q)$ also a valid transformation example. Similarly if the transformation is associative or composable and commutative.

Above strategies allow forming a diversed pool of embedding and transformation supervision for training. This can be further enhanced if it's easy to generate examples on the fly. For instance given a text domain example "a girl is sleeping on the beach", and transformation "replace word $a$ with word $b$", a lot of examples can be generated by picking different $a$ and $b$.

\section{Experiment on CSS Dataset}

\begin{figure}[t]
\begin{center}
\includegraphics[width=0.99\linewidth]{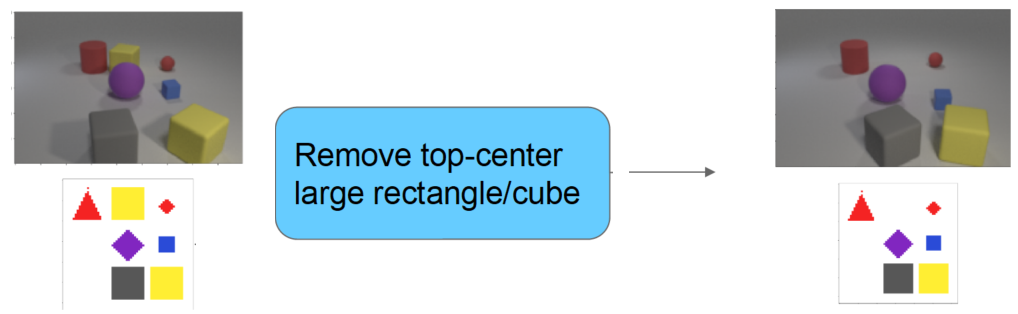}
\end{center}
   \caption{Example data of image retrieval with composition query in the CSS dataset.}
\label{fig:css}
\end{figure}

First we experiment transferring transformation between 2 different image domains on the synthesized dataset CSS \cite{our}. It was created for the task of image retrieval from image text compositional query; for each scene there is a 2D simple blobs and 3D rendered version image; an example query is shown in figure\ref{fig:css}. This naturally fits into our framework: (1) such composition query can be interpreted as a query image plus a transformation, here described by the text (2) we can train on the 2D images (source domain) and test on 3D images (target domain).  

\begin{figure}[t]
\begin{center}
\includegraphics[width=0.99\linewidth]{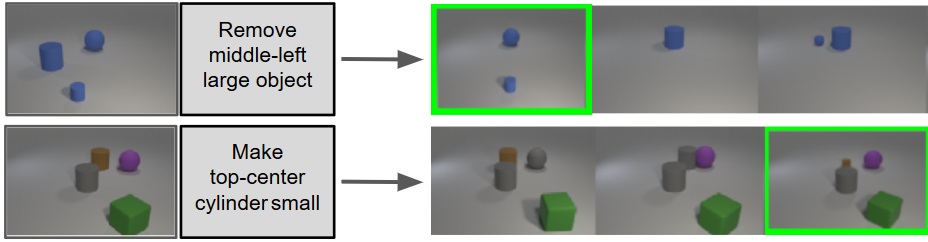}
\end{center}
   \caption{Example 3D-to-3D retrieval result on CSS Dataset}
\label{fig:css_qual}
\end{figure}

\subsection{Setup}

The dataset has ~16k (query, target) pairs for training. We use this as supervision for learning the transformation function using 2D images only. The image part of all queries come from a set of 1k base images. We use both 2D and 3D versions of these base images for learning the shared embedding between 2 domains. During test time, we will perform the same retrieval benchmark with the 3D image versions as in \cite{our}. 

Note that we are pretending that we don't have access to transformation examples of 3D images. We do have: (1) a lot of transformation examples in the 2D image domain (16k) and (2) a small amount of similarity labels between 2 domains (1k). This set up is to motivate our work: train on source domain and transfer to target domain where supervision is not available or more expensive (in fact \cite{our} states that generating all these 2D images only take minues while it's days for the 3D version even with a GPU).

\subsection{Implementation detail}

We used ResNet-18 to encode the 2D and 3D images, and LSTM for the transformation specification text; feature size for all of them is 512. The text feature is treated as transformation parameter. We train for 150k iterations with learning rate of 0.01. 

Set up and baselines are the same as \cite{our}, we train the same system, but without any transformation transferring, for 3 cases: transformation retrieval for 2D-to-2D images, 2D-to-3D and 3D-to-3D. The main experiment is the 3D-to-3D case where we can compare directly the baseline: learning the transformation in 3D image feature space, versus our approach: learning the transformation in 2D image feature space and share it.

\subsection{Result}

\begin{figure*}
\begin{center}
\includegraphics[width=0.9\linewidth]{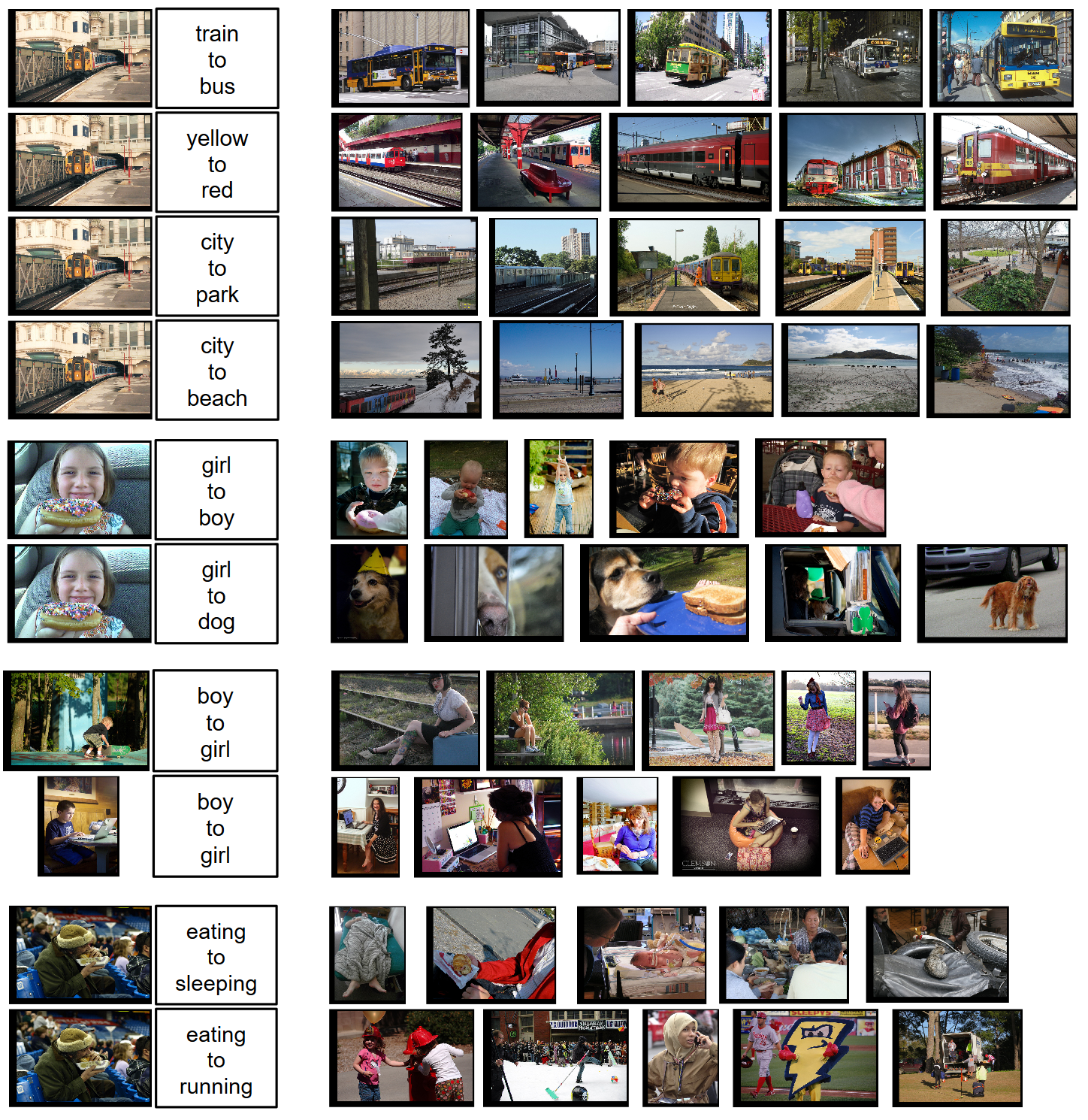}
\end{center}
   \caption{Example retrieval result on COCO-val2014}
\label{fig:cocoqual}
\end{figure*}

We reported R@1 performance in Table \ref{tab:css}, some qualitative result is shown in Figure \ref{fig:css_qual}. Our approach, without 3D-to-3D training examples, achieves comparable result to training on 3D-to-3D transformation examples. Transferring appears to be very effective for this simple dataset.

Since our method learns a shared embedding, we can do cross domain retrieval. In the 2D-to-3D retrieval case, surprisingly ours outperform actually training on 2D-to-3D examples baseline. This suggests learning a cross domain transformation retrieval is more challenging than learning in-domain and then share.

\section{Experiment on Transferring Text Transformation to Image}

While composing and manipulating texts is everyday task to people (who can write and read), composing and manipulating images are specialized skills. In this section we attempt to transfer text transformation to images. Note that there's inherently differences between the vision and language, something can be described in one domain but might be difficult to fully translate to other domain. In \cite{young2014image}, the denotation graph is introduced, where each node represents a text expression and is grounded with a set of example images; each edge in the graph represents a text transformation.

\subsection{Word replacement text transformation}

\begin{figure}[t]
\begin{center}
\includegraphics[width=0.9\linewidth]{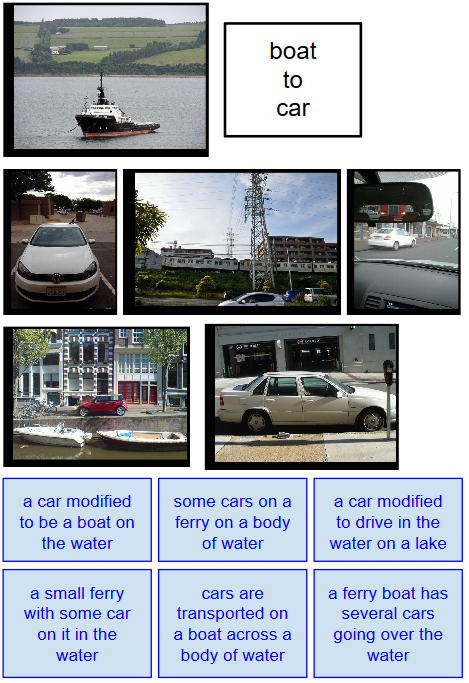}
\end{center}
   \caption{An example retrieval result: top is the query, bottom is retrieved images and captions.}
\label{fig:cocoqual4}
\end{figure}

We choose a very simple transformation to study here: given a text, replace a particular word with another word. As in previous experiment, we use image retrieval as demonstration task. For example if applying a text transformation of [replace beach with street] to the image of a dog running on the beach,  we would want retrieve images of dogs running on the street (2nd example in figure 1).

However, exact expectation of the result is hard to define especially if the image is crowded with different things (what street scene is desired, should other objects in the scene be preserved, should the dog be kept at exact pose, or at instance level or at category level, etc). In addition to such ambiguousness, composing images is not trivial, hence collecting labels is very difficult. One can explicitly define a specific transformation in the image domain, then equate it to another specific transformation in the text domain through machine learning; while interesting, it's not what we want study here. Our approach allows training a transformation in one domain and transfer it to the other domain without any transformation examples in the target domain.

\subsection{Setup}

We use the COCO train2014 dataset \cite{lin2014microsoft} to learn the join embedding of images and texts; it has around 80k images, each is accompanied with 5 captions.

We create a list of hundred of word replacement pairs from a pool of around 100 words (for example "beach to street", "boy to girl", "dogs to people", etc); theses words are manually chosen such that they are popular and visually distinctive concepts. During training, we apply word replacement to the captions to generate transformation examples on the fly.

We used pretrained ResNet-50 for encoding images, not fine-tuning the conv layers, and LSTM for encoding captions and words; the embedding size is 512. 
The parameter $t$ for the transformation is the word replacement pair, $\hat{t}$ will be the concatenation of the encoded representations of the word to be replaced and the new word. As defined in section 3, the transformation function can be a simple 2 layers feed forward network, or a recent technique TIRG \cite{our}.

Note that since this word replacement transformation is reversible and generating new text examples on the fly is easy, we take advantage of data augmentation tricks mentioned in section 3.3.

\subsection{Qualitative Result on COCO}

As mentioned in section 5.1, correct retrieval result can be ambiguous to define, so we mainly focus on demonstration of qualitative result here. The COCO-val2014 with around 40k images will be used as retrieval database.

We show some result in figure \ref{fig:cocoqual}. Somewhat reasonable retrieval result can be obtained if the replacing words are really popular and visually distinctive; this includes COCO object categories (person, dog, car, ...) or common scenes (room, beach, park, ...). To us, a reasonable result would be the introduction of the concept represented by the new word, while other elements in the image (subjects, background, composition,...) are kept unchanged as much as possible.

Replacing adjectives and verbs are difficult. Popular adjectives are object like attribute such as "woody" or "grassy". Abstract adjectives are rare in this dataset, some might be "young", "small" or "big". Colors might be ones that have better chance of working in our experience since they are often visually distinctive.

Verbs are the most challenging (for example last rows in figure \ref{fig:cocoqual}). We speculate the system relies less on verbs to matching image and text since nouns/objects are informative enough and easier to learn from (for context, recent research have demonstrated object recognition performance at superhuman level, but action recognition remains challenging). Also it could be partly because COCO is object oriented, so is ImageNet which is used for pretraining.

Finally, we note that there's still discrepancy between images and texts even in a shared embedding. In figure \ref{fig:cocoqual4}, we show an example where top ranked images and texts are retrieved but they are not reflecting the same semantics. Hence our task could benefit with improvement in image text matching method (the one we use in this work is basic and straightforward, but slightly inferior to state of the art approaches on image-to-text and text-to-image retrieval benchmarks).

\subsection{Simple Image Captions (SIC112) benchmark}

\begin{figure}[t]
\begin{center}
\includegraphics[width=0.9\linewidth]{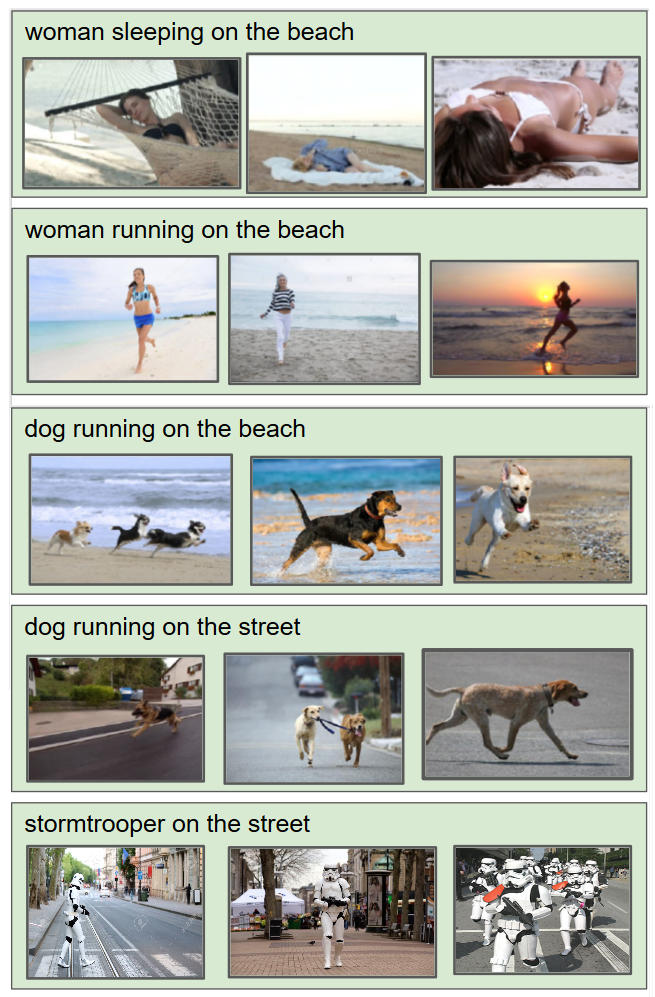}
\end{center}
   \caption{The SIC112 dataset}
\label{fig:nams_dataset}
\end{figure}

In order to have a quantitative measurement, we collect a small benchmark for this task. First we manually define a set of 112 very simple, attribute-like captions, each can contain a subject (man, woman, dog, etc), a verb (running, sleeping, sitting) and a background (on the street, beach, ect). For each caption we perform a google image search to collect images, then manually filter them. On average we collect 15 images for each caption. We call this the Simple Image Captions 112 (SIC112) dataset, some examples are shown in Figure \ref{fig:nams_dataset}.

With this dataset, we can now test our image retrieval task quantitatively, by using the captions as the label for images. The retrieval is considered success if the retrieved image has the same caption label corresponding to the query image's caption after applying the word replacement. We use the Recall at rank k (R@k) metric, which is defined as the percentage of test cases in which top k retrieved result contains at least 1 correct image. Note that the dataset is for testing only, training is done on COCO-train2014 as described in previous section.

\textbf{Baselines}: we compare with the following
\begin{enumerate}
\item Image Only: ignore the transformation and do image to image retrieval.
\item Embedding arithmetic: the word replacing transformation can be done by directly adding and subtracting their corresponding embedding. This simple strategy has been found to be effective in previous works on text embedding \cite{mikolov2013distributed} (for example ``king" - ``man" + ``woman" = ``queen"), image synthesis \cite{radford2015unsupervised} and 3D model and  text joint embedding \cite{chen2018text2shape}. 
\item Image to Text to Image retrieval: instead of transferring the transformation, this baseline translates the query image to text, perform the transformation natively and then translate it back to image. Here the translation is done by the our image text matching system since it is capable of retrieval in both direction (image to text and text to image). For image to text, our implementation uses COCO-train2014 dataset of ~400k captions as text database; an alternative could be an out of the box image captioning system.
\item Text (Ground truth target caption) to Image retrieval: this is similar to the last baseline, assuming a perfect image to text translation is given, so the ground truth caption will be used as query for retrieval.
\end{enumerate}

\begin{table*}
\begin{center}
\begin{tabular}{|l|cccc|ccc|}
\hline
Test queries & \multicolumn{3}{c} {Keep others 2, only change:} & keep novel subject, & \multicolumn{3}{c|} {All}   \\
- & subject & verb & background & change background & R@1 & R@5 & R@10  \\
\hline\hline
Image Only & 0 & 0 & 0 & 0 & 0 & 12.9$^{\pm0.1}$ & 21.9$^{\pm0.2}$ \\
Arithmetic & 19.2$^{\pm1.1}$ & 12.5$^{\pm2.1}$ & \underline{23.4}$^{\pm1.1}$ & \textbf{43.5}$^{\pm2.5}$ & \underline{20.5}$^{\pm0.6}$ & \underline{59.5}$^{\pm0.6}$ & \underline{75.0}$^{\pm0.4}$ \\
Img2Text2Img & \underline{19.7}$^{\pm0.6}$ & \underline{14.9}$^{\pm0.7}$ & 17.0$^{\pm0.6}$ & 12.5$^{\pm2.5}$ & 17.6$^{\pm0.4}$ & 45.7$^{\pm0.2}$ & 59.9$^{\pm0.2}$ \\
Our & \textbf{27.4}$^{\pm0.9}$ & \textbf{23.5}$^{\pm0.6}$ & \textbf{26.0}$^{\pm0.9}$ & \underline{38.2}$^{\pm2.6}$ & \textbf{26.6}$^{\pm0.4}$ & \textbf{64.0}$^{\pm0.4}$ & \textbf{78.3}$^{\pm0.6}$ \\
\hline 
GT caption & 40.6$^{\pm3.5}$ & 40.3$^{\pm3.0}$ & 39.8$^{\pm3.2}$ & 29.6$^{\pm9.0}$ & 39.6$^{\pm3.3}$ & 82.4$^{\pm3.0}$ & 92.0$^{\pm1.8}$ \\
\hline
\end{tabular}
\end{center}
\caption{R@1 retrieval result on the SIC112 benchmark.}
\label{tab:test}
\end{table*}

\begin{figure*}
\begin{center}
\includegraphics[width=0.9\linewidth]{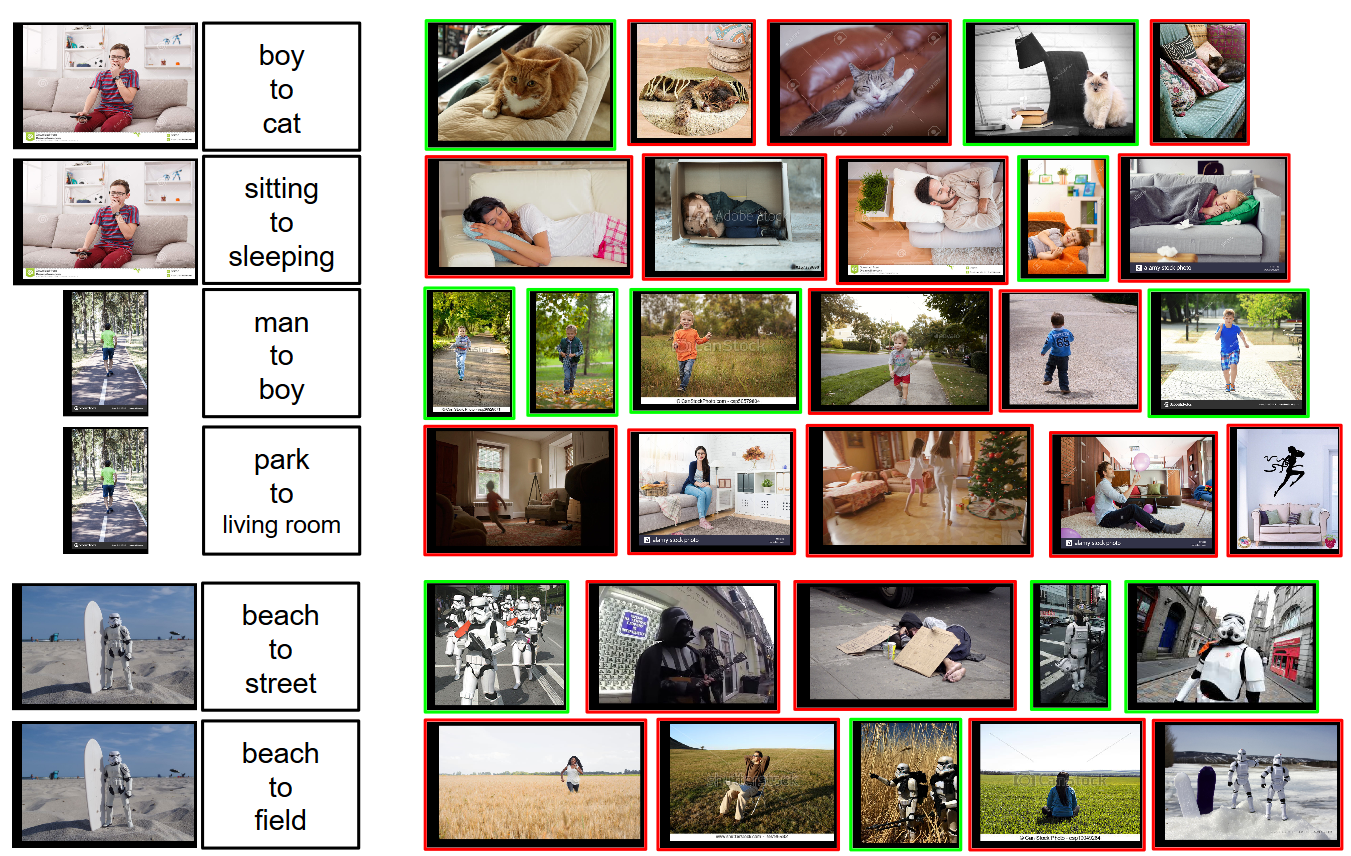}
\end{center}
   \caption{Example retrieval result on SIC112 benchmark.}
\label{fig:namsdatasetqual}
\end{figure*}

\textbf{Result}:
Some qualitative result is shown in figure \ref{fig:namsdatasetqual}. The retrieval performance is reported in table \ref{tab:test}. For analysis we split test queries into 4 groups: 8205 queries changing subject (for example ``girl to boy"), 3282 changing verb, 6564 changing background, and 745 special queries changing background of images which contain novel subjects (such as ``stormtrooper" that doesn't appear in COCO2014train). The last group is to demonstrate the use case where direct translation between image and text might be difficult and transferring might be more appropriate. 

We consider the GT caption to image retrieval baseline as the upper bound. Among subject, verb and background, changing verb seems more challenging. Our approach outperforms the other baselines demonstrating it is more beneficial to perform the transformation in the same image domain than translating to text. Still ours is much worse than GT caption baseline, suggesting there is still a lot of room for improvement. In particular our approach could benefit a lot from better image-text joint embedding technique.

On keep novel subject change background queries, translating to text or even using GT caption result in worse performance because the system can not recognize the novel object in text domain. Performing transformation in the native image domain by embedding arithmetic operation or our approach fits this use case better. Arithmetic baseline performs very well on changing background, and even outperform ours when verb is not involved. This baseline also has the advantage that it's simple and no additional learning need to be done. However we'd expect when the operation is more complex or subtle (for example keep verb unchanged, or change verb, or dealing with more complex captions like in COCO2014), learning a transformation function would be better than relying on simple arithmetic operation.

\section{Conclusion}

We propose to learn a feature transformation function where no training examples are available by learning such function on another domain with similar semantics (where training examples are abundant). Then it can be transfer to the original target domain by shared embedding feature space. We demonstrate such transformed feature can be very useful in image retrieval application. One can also learn a decoding function, for example, to generate image or text from the feature. Future works could study more complex text transformation and semantic composition beyond simple ``word replacing".

{\small
\bibliographystyle{ieee}
\bibliography{egbib}

\begin{thebibliography}{10}\itemsep=-1pt

\bibitem{ak2018learning}
K.~E. Ak, A.~A. Kassim, J.~H. Lim, and J.~Y. Tham.
\newblock Learning attribute representations with localization for flexible
  fashion search.
\newblock In {\em CVPR}, 2018.

\bibitem{akata2013label}
Z.~Akata, F.~Perronnin, Z.~Harchaoui, and C.~Schmid.
\newblock Label-embedding for attribute-based classification.
\newblock In {\em Proceedings of the IEEE Conference on Computer Vision and
  Pattern Recognition}, pages 819--826, 2013.

\bibitem{chen2018text2shape}
K.~Chen, C.~B. Choy, M.~Savva, A.~X. Chang, T.~Funkhouser, and S.~Savarese.
\newblock Text2shape: Generating shapes from natural language by learning joint
  embeddings.
\newblock {\em arXiv preprint arXiv:1803.08495}, 2018.

\bibitem{choi2017stargan}
Y.~Choi, M.~Choi, M.~Kim, J.-W. Ha, S.~Kim, and J.~Choo.
\newblock Stargan: Unified generative adversarial networks for multi-domain
  image-to-image translation.
\newblock {\em arXiv preprint arXiv:1711.09020}, 2017.

\bibitem{cordts2016cityscapes}
M.~Cordts, M.~Omran, S.~Ramos, T.~Rehfeld, M.~Enzweiler, R.~Benenson,
  U.~Franke, S.~Roth, and B.~Schiele.
\newblock The cityscapes dataset for semantic urban scene understanding.
\newblock In {\em Proceedings of the IEEE conference on computer vision and
  pattern recognition}, pages 3213--3223, 2016.

\bibitem{deng2014large}
J.~Deng, N.~Ding, Y.~Jia, A.~Frome, K.~Murphy, S.~Bengio, Y.~Li, H.~Neven, and
  H.~Adam.
\newblock Large-scale object classification using label relation graphs.
\newblock In {\em European conference on computer vision}, pages 48--64.
  Springer, 2014.

\bibitem{frome2013devise}
A.~Frome, G.~S. Corrado, J.~Shlens, S.~Bengio, J.~Dean, T.~Mikolov, et~al.
\newblock Devise: A deep visual-semantic embedding model.
\newblock In {\em Advances in neural information processing systems}, pages
  2121--2129, 2013.

\bibitem{guo2018dialog}
X.~Guo, H.~Wu, Y.~Cheng, S.~Rennie, and R.~S. Feris.
\newblock Dialog-based interactive image retrieval.
\newblock {\em arXiv preprint arXiv:1805.00145}, 2018.

\bibitem{han2017automatic}
X.~Han, Z.~Wu, P.~X. Huang, X.~Zhang, M.~Zhu, Y.~Li, Y.~Zhao, and L.~S. Davis.
\newblock Automatic spatially-aware fashion concept discovery.
\newblock In {\em ICCV}, 2017.

\bibitem{isola2017image}
P.~Isola, J.-Y. Zhu, T.~Zhou, and A.~A. Efros.
\newblock Image-to-image translation with conditional adversarial networks.
\newblock In {\em 2017 IEEE Conference on Computer Vision and Pattern
  Recognition (CVPR)}, pages 5967--5976. IEEE, 2017.

\bibitem{jiang2012leveraging}
L.~Jiang, A.~G. Hauptmann, and G.~Xiang.
\newblock Leveraging high-level and low-level features for multimedia event
  detection.
\newblock In {\em ACM MM}, 2012.

\bibitem{johnson2018image}
J.~Johnson, A.~Gupta, and L.~Fei-Fei.
\newblock Image generation from scene graphs.
\newblock {\em arXiv preprint arXiv:1804.01622}, 2018.

\bibitem{kovashka2012whittlesearch}
A.~Kovashka, D.~Parikh, and K.~Grauman.
\newblock Whittlesearch: Image search with relative attribute feedback.
\newblock In {\em Computer Vision and Pattern Recognition (CVPR), 2012 IEEE
  Conference on}, pages 2973--2980. IEEE, 2012.

\bibitem{lampert2009learning}
C.~H. Lampert, H.~Nickisch, and S.~Harmeling.
\newblock Learning to detect unseen object classes by between-class attribute
  transfer.
\newblock In {\em Computer Vision and Pattern Recognition, 2009. CVPR 2009.
  IEEE Conference on}, pages 951--958. IEEE, 2009.

\bibitem{lin2014microsoft}
T.-Y. Lin, M.~Maire, S.~Belongie, J.~Hays, P.~Perona, D.~Ramanan,
  P.~Doll{\'a}r, and C.~L. Zitnick.
\newblock Microsoft coco: Common objects in context.
\newblock In {\em European conference on computer vision}, pages 740--755.
  Springer, 2014.

\bibitem{mai2017spatial}
L.~Mai, H.~Jin, Z.~Lin, C.~Fang, J.~Brandt, and F.~Liu.
\newblock Spatial-semantic image search by visual feature synthesis.
\newblock In {\em Computer Vision and Pattern Recognition (CVPR), 2017 IEEE
  Conference on}, pages 1121--1130. IEEE, 2017.

\bibitem{mikolov2013distributed}
T.~Mikolov, I.~Sutskever, K.~Chen, G.~S. Corrado, and J.~Dean.
\newblock Distributed representations of words and phrases and their
  compositionality.
\newblock In {\em Advances in neural information processing systems}, pages
  3111--3119, 2013.

\bibitem{radford2015unsupervised}
A.~Radford, L.~Metz, and S.~Chintala.
\newblock Unsupervised representation learning with deep convolutional
  generative adversarial networks.
\newblock {\em arXiv preprint arXiv:1511.06434}, 2015.

\bibitem{raj2018swapnet}
A.~Raj, P.~Sangkloy, H.~Chang, J.~Hays, D.~Ceylan, and J.~Lu.
\newblock Swapnet: Image based garment transfer.

\bibitem{reed2016learning}
S.~Reed, Z.~Akata, H.~Lee, and B.~Schiele.
\newblock Learning deep representations of fine-grained visual descriptions.
\newblock In {\em Proceedings of the IEEE Conference on Computer Vision and
  Pattern Recognition}, pages 49--58, 2016.

\bibitem{reed2016generative}
S.~Reed, Z.~Akata, X.~Yan, L.~Logeswaran, B.~Schiele, and H.~Lee.
\newblock Generative adversarial text to image synthesis.
\newblock {\em arXiv preprint arXiv:1605.05396}, 2016.

\bibitem{ren2014face}
S.~Ren, X.~Cao, Y.~Wei, and J.~Sun.
\newblock Face alignment at 3000 fps via regressing local binary features.
\newblock In {\em Proceedings of the IEEE Conference on Computer Vision and
  Pattern Recognition}, pages 1685--1692, 2014.

\bibitem{richter2016playing}
S.~R. Richter, V.~Vineet, S.~Roth, and V.~Koltun.
\newblock Playing for data: Ground truth from computer games.
\newblock In {\em European Conference on Computer Vision}, pages 102--118.
  Springer, 2016.

\bibitem{rui1998relevance}
Y.~Rui, T.~S. Huang, M.~Ortega, and S.~Mehrotra.
\newblock Relevance feedback: a power tool for interactive content-based image
  retrieval.
\newblock {\em IEEE Transactions on circuits and systems for video technology},
  8(5):644--655, 1998.

\bibitem{sangkloy2016sketchy}
P.~Sangkloy, N.~Burnell, C.~Ham, and J.~Hays.
\newblock The sketchy database: learning to retrieve badly drawn bunnies.
\newblock {\em ACM Transactions on Graphics (TOG)}, 35(4):119, 2016.

\bibitem{sangkloy2016scribbler}
P.~Sangkloy, J.~Lu, C.~Fang, F.~Yu, and J.~Hays.
\newblock Scribbler: Controlling deep image synthesis with sketch and color.
\newblock {\em Computer Vision and Pattern Recognition, CVPR}, 2017.

\bibitem{shrivastava2017learning}
A.~Shrivastava, T.~Pfister, O.~Tuzel, J.~Susskind, W.~Wang, and R.~Webb.
\newblock Learning from simulated and unsupervised images through adversarial
  training.
\newblock In {\em Proceedings of the IEEE Conference on Computer Vision and
  Pattern Recognition}, pages 2107--2116, 2017.

\bibitem{tzeng2017adversarial}
E.~Tzeng, J.~Hoffman, K.~Saenko, and T.~Darrell.
\newblock Adversarial discriminative domain adaptation.
\newblock In {\em 2017 IEEE Conference on Computer Vision and Pattern
  Recognition (CVPR)}, pages 2962--2971. IEEE, 2017.

\bibitem{vo2018generalization}
N.~Vo and J.~Hays.
\newblock Generalization in metric learning: Should the embedding layer be the
  embedding layer?
\newblock {\em arXiv preprint arXiv:1803.03310}, 2018.

\bibitem{our}
N.~Vo, L.~Jiang, C.~Sun, K.~Murphy, L.-J. Li, L.~Fei-Fei, and J.~Hays.
\newblock Composing text and image for image retrieval - an empirical odyssey.
\newblock In {\em CVPR}, 2019.

\bibitem{wang2014learning}
J.~Wang, Y.~Song, T.~Leung, C.~Rosenberg, J.~Wang, J.~Philbin, B.~Chen, and
  Y.~Wu.
\newblock Learning fine-grained image similarity with deep ranking.
\newblock In {\em CVPR}, 2014.

\bibitem{wang2016learning}
L.~Wang, Y.~Li, and S.~Lazebnik.
\newblock Learning deep structure-preserving image-text embeddings.
\newblock In {\em CVPR}, 2016.

\bibitem{weinberger2009distance}
K.~Q. Weinberger and L.~K. Saul.
\newblock Distance metric learning for large margin nearest neighbor
  classification.
\newblock {\em Journal of Machine Learning Research}, 10(Feb):207--244, 2009.

\bibitem{xian2016texturegan}
V.~A. A. R. J. L. C. F. F. Y. J.~H. Wenqi~Xian, Patsorn~Sangkloy.
\newblock Texturegan: Controlling deep image synthesis with texture patches.
\newblock {\em Computer Vision and Pattern Recognition, CVPR}, 2018.

\bibitem{xu2017scene}
D.~Xu, Y.~Zhu, C.~B. Choy, and L.~Fei-Fei.
\newblock Scene graph generation by iterative message passing.
\newblock {\em arXiv preprint arXiv:1701.02426}, 2017.

\bibitem{young2014image}
P.~Young, A.~Lai, M.~Hodosh, and J.~Hockenmaier.
\newblock From image descriptions to visual denotations: New similarity metrics
  for semantic inference over event descriptions.
\newblock {\em Transactions of the Association for Computational Linguistics},
  2:67--78, 2014.

\bibitem{zhao2017memory}
B.~Zhao, J.~Feng, X.~Wu, and S.~Yan.
\newblock Memory-augmented attribute manipulation networks for interactive
  fashion search.
\newblock In {\em Computer Vision and Pattern Recognition (CVPR), 2017 IEEE
  Conference on}, pages 6156--6164. IEEE, 2017.

\end{thebibliography}
}

\end{document}